\documentclass[conference]{IEEEtran}
\IEEEoverridecommandlockouts

\usepackage{cite}
\usepackage{amsmath,amssymb,amsfonts}
\usepackage{algorithmic}
\usepackage{graphicx}
\usepackage{textcomp}
\usepackage{xcolor}
\def\BibTeX{{\rm B\kern-.05em{\sc i\kern-.025em b}\kern-.08em
    T\kern-.1667em\lower.7ex\hbox{E}\kern-.125emX}}
\begin{document}

\title{When End-to-End is Overkill: \\
Rethinking Cascaded Speech-to-Text Translation
}

\author{\IEEEauthorblockN{1\textsuperscript{st} Anna Min}
\IEEEauthorblockA{\textit{School of Software} \\
\textit{Tsinghua University}\\
Beijing, China \\
man20@mails.tsinghua.edu.cn}
\and
\IEEEauthorblockN{2\textsuperscript{nd} Chenxu Hu}
\IEEEauthorblockA{\textit{IIIS} \\
\textit{Tsinghua University}\\
Beijing, China \\
hu-cx21@mails.tsinghua.edu.cn}
\and
\IEEEauthorblockN{3\textsuperscript{rd} Yi Ren}
\IEEEauthorblockA{
\textit{TickTok}\\
ren.yi@bytedance.com}
\and
\IEEEauthorblockN{4\textsuperscript{th} Hang Zhao}
\IEEEauthorblockA{\textit{IIIS} \\
\textit{Tsinghua University}\\
Beijing, China \\
hangzhao@tsinghua.edu.cn}

}

\maketitle

\begin{abstract}
Though end-to-end speech-to-text translation has been a great success, we argue that the cascaded speech-to-text translation model still has its place, which is usually criticized for the error propagation between automatic speech recognition (ASR) and machine translation (MT) models. In this paper, we explore the benefits of incorporating multiple candidates from ASR and self-supervised speech features into MT. Our analysis reveals that the primary cause of cascading errors stems from the increased divergence between similar samples in the speech domain when mapped to the text domain. By including multiple candidates and self-supervised speech features, our approach allows the machine translation model to choose the right words and ensure precise translation using various speech samples. This strategy minimizes error spread and takes advantage of large ASR and MT datasets, along with pre-trained ASR/MT models, while addressing associated issues.
\end{abstract}


\section{Introduction}

In recent years, the academic community has been intrigued by the rapid advancement of end-to-end speech-to-text translation models~\cite{berard2016listen}. These efficient encoder-decoder architectures provide a direct avenue for translating speech, bypassing the need for complex intermediate symbolic representations. However, the arduous task of assembling and curating end-to-end data poses a significant challenge, entailing considerable costs and extensive efforts. These end-to-end methods need the careful selection of high-quality data or argumentation~\cite{popuri2022enhanced,jia2019leveraging,stoian2020analyzing,pino2020self,wang2021large}, encompassing both speech and translated transcripts, and the scrupulous exclusion of erroneous examples.

Nonetheless, cascaded speech-to-text translation models have encountered substantial criticism due to an intrinsic shortcoming called “cascaded loss” or “error propagation”. Studies on ASR+MT systems have explored various methods to enhance the integration of ASR output lattices into MT models~\cite{matusov2005integration,quan2005integrated,ma2020neural}. To mitigate error propagation, several approaches~\cite{bertoldi2005new,beck2019neural,sperber2019self,peitz2012spoken,cheng2019breaking,di2019robust,dalmia2021searchable,inaguma2021fast} have been proposed to integrate ASR and MT models for end-to-end models, which necessitate the addition of supplementary modules and substantial additional training. In contrast, the method proposed in our research employs an n-best strategy that does not require additional parameters and can significantly enhance performance with minimal fine-tuning.


\begin{figure}[ht]
  \centering
    \centering
    \includegraphics[width=\linewidth]{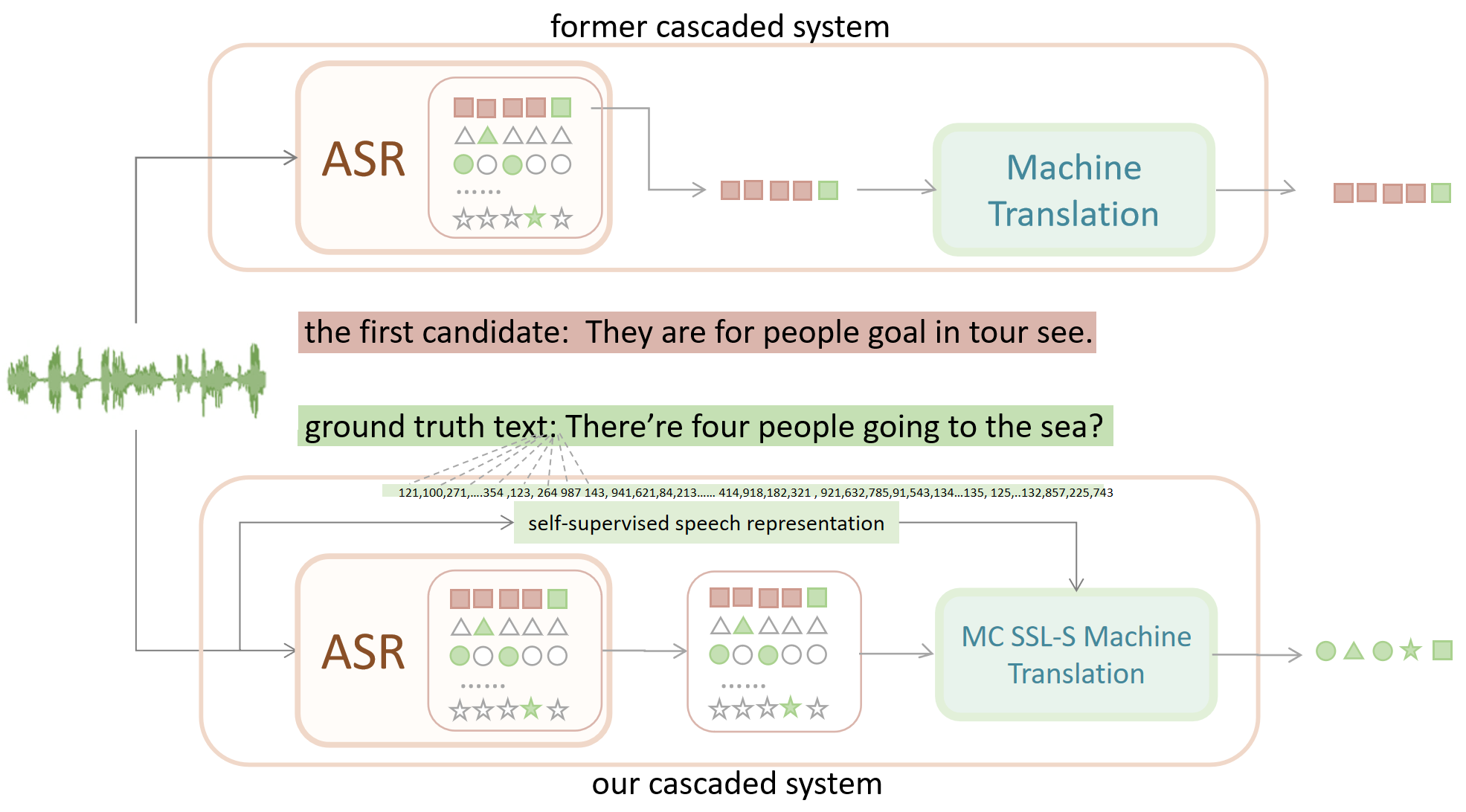}
\caption{The correct words are scattered among various candidates, while the former cascaded system directly selected the first candidate, resulting in similar pronunciation errors in the ASR output that are further propagated through the translation model, causing cascading losses.}
\end{figure}


Recent studies~\cite{bommasani2021opportunities,han2021pre} have demonstrated the performance improvements achieved by scaling up pre-trained models for downstream natural language processing tasks. To fully exploit the potential of pre-trained MT and ASR models, we present a novel perspective on error propagation and the preservation of essential speech information. We propose the idea of utilizing multiple ASR candidates for machine translation, integrated with self-supervised speech representations, to enhance the accuracy of the translation. Our comprehensive analysis reveals the primary causes of error propagation in cascaded systems, which originate from the misalignment between the acoustic and semantic dimensions of speech. Factors such as homophones with different meanings and word elisions contribute to inaccuracies in ASR results, which consequently propagate to the machine translation model. Furthermore, we explore the use of self-supervised language representations to preserve fine-grained linguistic information in speech.  

Our model has the following advantages:\vspace{-2pt} 

\begin{enumerate}
    \item Our model achieves the best performance among cascaded models in the Speech-to-Text(S2T) translation task.

    \item Our method can leverage variously known ASR and machine translation (MT) pre-trained models. Specifically, it can effortlessly adapt to different model architectures without the need to adjust the model parameters. 

    \item Unlike end-to-end models, our approach does not require an extensive large amount of costly \textless speech, transcript, target\textgreater     paired data. 

    \item Our model demonstrates rapid training speed and attains exceptional results with minimal data utilization for fine-tuning the Machine Translation (MT) model. 
\end{enumerate}

\section{Analysis}

To see where the error propagation lies in the cascaded system and how ASR errors propagate to MT, we pose two questions: 1) Can the top-ranked ASR candidate cover all the lexicons? 2) Is the top-ranked ASR candidate always the best translation result?

\subsection{Preliminary experiments}

In the cascaded system, we extract the top 20 results based on the scores from the ASR system and then select the top $n$ candidates. We calculate the lexical overlap between these candidates $\{c_1, c_2 ... c_n\}$ and the ground truth text $\{gt\}$ of the ASR, which means the length of set of all words in candidates $\{ w| w \in c_k, 1 \leq k \leq n\}$ intersected with $\{ w| w \in gt\}$ divided by the length of the latter. In Table 1, “average” refers to the average lexical overlap between each candidate and the ground truth. At the same time “cumulative” represents the lexical overlap when considering a combination of $n$ candidates with the ground truth. We observe that as $n$ increases, although the average lexical overlap decreases, the cumulative overlap improves. This indicates that apart from the top-ranked candidate, the ASR system fails to include some vocabulary, which remains in the lower-ranked candidates.

\begin{table}[th]
  \caption{Lexical overlap between ASR candidates and GT}
  \centering
  \begin{tabular}{ |c|c|c| }
    \hline
    \multicolumn{1}{|c}{\textbf{$n$ candidates}} & 
    \multicolumn{1}{|c|}{\textbf{Average}}  & 
    \multicolumn{1}{c|}{\textbf{Cumulative}} \\
    \hline
1                 & 92.0\% & 92.0\%  \\
5                 & 90.0\% & 94.3\%  \\ 
10                 & 89.4\% & 95.0\%  \\ 
20                 & 89.5\% & 95.5\% \\
    \hline
  \end{tabular}
\end{table}

\vspace{-0.2cm}

Additionally, we use a trained MT model to translate and calculate BLEU scores for the top 5 candidates based on ASR scores. We then analyze the index and percentage of the highest BLEU score and find that only 45.35\% of the candidates with the highest BLEU score corresponded to the candidate with the lowest word error rate. It indicates that the top-ranked ASR candidate always performs the best translation result. 

We also compare the BLEU score of the translation for the candidate with the lowest word error rate with the average BLEU score of the top 5 translations. The BLEU result of the translated best ASR candidate is 36.0, while the BLEU result of the translated first ASR candidate is 32.9. This indicates that combining multiple candidates can significantly improve translation results.

\begin{table}[th]
  \caption{The proportion of the ASR result index of the best candidate based on the BLEU score of the translation result} 
  \centering
  \begin{tabular}{|c|ccccc|}
    \hline
    {Best BLEU idx}  & 1  & 2 & 3 & 4 & 5  \\ 
    \hline
    {percentage(\%)}  & 45.35 & 16.73  & 14.07 & 12.12 & 11.74 \\
    \hline
  \end{tabular}
\end{table}


\vspace{-0.6cm}

\subsection{Source of cascade loss}




Since the best result in the ASR stage does not necessarily represent the best result in an end-to-end manner, there must be a point where cascaded errors occur. Therefore, we delve deeper into analyzing these cascaded errors and make the hypothesis that the source of cascaded loss mainly arises from discrepancies between the pronunciation space and semantic space and that ASR models encounter difficulties in selecting results that most accurately match the language patterns. 

ASR is trained using paired speech and text data, but the language patterns it captures are not as rich as those in translation models. Therefore, when the recognition results are similar, it is difficult for the ASR model to select the candidate that best aligns with human common sense and grammar based on scores alone.

As an example, consider the phrase “has put the race on the top.” The highest-scoring candidate in ASR recognizes “race” as “rays,” which contradicts common sense. However, subsequent candidates include various results with similar pronunciations, such as “race,” “raised,” and “raise.” Another example is “Recording the transaction in an immutable distributed ledger,” while the highest-scoring result in ASR is “Recording the transaction in an immutable distributed lecture,” which is not a common expression. The subsequent candidates include “legend,” “literature,” “letter,” “ledger,” and other results.


\vspace{-0.35cm}

\begin{figure}[ht]
  \centering
    \centering
    \includegraphics[width=0.9\linewidth]{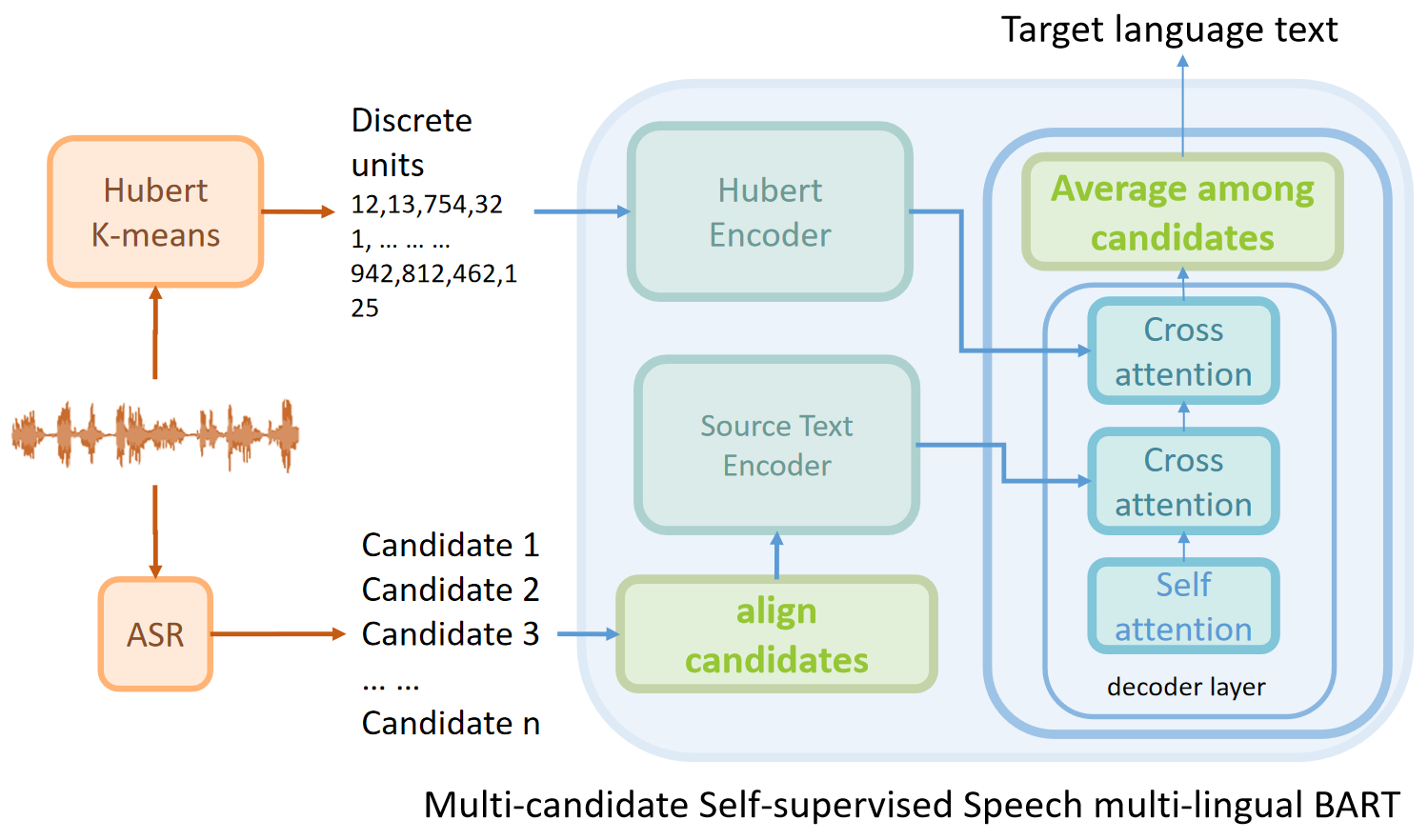}
    \caption{Overview of our proposed MC-sslS system}
\end{figure}

\section{Method}

To leverage the powerful capability of machine translation in capturing semantic patterns, we propose utilizing multi-candidate ASR inputs and averaging attention computation in the MT model. Furthermore, to address the issue of error propagation in ASR caused by homophones, we employ self-supervised speech representations to enhance accuracy. The combination of methods we propose is represented by the model depicted in Figure 2.

\subsection{Multi-candidate from ASR}

The correspondence between the speech domain and the semantic domain is not always perfect. However, compared to machine translation models, ASR models have limited ability to capture semantic patterns effectively. Consequently, ASR is prone to mapping speech samples that are acoustically similar to semantically distant outputs. Moreover, ASR cannot selectively choose samples that align more closely with human language conventions and patterns. For a given speech input, the correct vocabulary might be dispersed among multiple candidates generated during beam search. Unfortunately, previous cascaded systems only consider the top-scoring candidate and pass it to the machine translation model, resulting in error propagation within the cascaded system.
Our cascaded model first uses Wenet~\cite{yao2021wenet} to perform ASR on GigaST~\cite{ye2022gigast}; we store the top $20$ ASR result sentences based on the cumulative log probability value from the end of beam search. 

\subsubsection{Aligned by common substrings} 
We follow the following steps to align the candidates:
\begin{enumerate}
    \item The top $n$ ranked texts $t_k(1 \leq k \leq n)$ consisting of $n_k$ words ($w_1,w_2....w_{n_k}$) are selected (in our following experiments, we set $n$ to be 5).
    \item The dynamic programming algorithm for finding the longest common substrings ~\cite{charalampopoulos_et_al:LIPIcs.ESA.2021.30} is used for $n-1$ times, in the preprocessing stage before putting into the model. There are $n-1$ processes in total. We denote $t_i^{m+1}$ as the aligned result of $t_i$ after the $m^{th}(1 \leq m \leq n-1)$ process. $t_1^{1}$ is equal to $t_1$.
    \item During the $m^{th}$ process, (1) we calculate the longest common subsequences of $\{t_1^{m},t_{m+1}\}$, and get $\{t_1^{m+1},t_{m+1}^{m+1}\}$ by connecting the substrings and the largest length of uncommon substrings. (2) The common substrings are aligned, and the remaining parts are padded with “unk” tokens. (3) When $2 \leq m \leq n-1$, the $\{t_2^{m}...t_{m+1}^{m}\}$ are padded with “unk” tokens at the same indexes where $\{t_1^{m}\}$ is padded to $\{t_1^{m+1}\}$. Then, one process ends. 
    \item  After $n-1$ processes, the aligned and padded texts of the same length are used as input to the attention-based machine translation model. 
\end{enumerate}
\vspace{-0.2cm}
\begin{figure}[ht]
  \centering
    \centering
    \includegraphics[width=\linewidth]{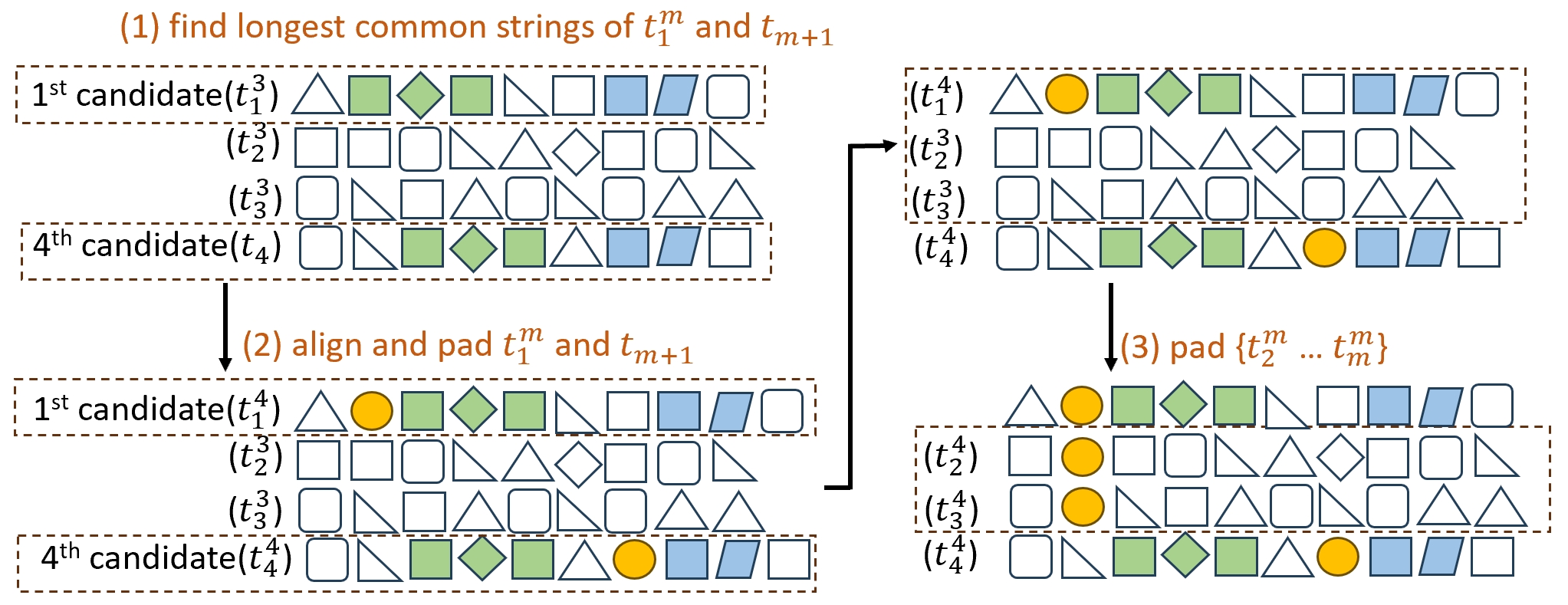}
    \caption{Above is how the $3^{rd}$ process is calculated. After finding the longest common subsequences, candidates are aligned and padded. The orange circles denote “unk” tokens.}
\end{figure}

\vspace{-0.2cm}

\subsubsection{Average attention among candidates} Thus, we propose an innovative approach, incorporating multiple ASR candidates into a single attention-based machine translation model. Here we use mBART as the backbone architecture. The multi-candidate model can share identical parameters with the translation model, differing only in its attention-averaging technique. 

During the training stage, this method averages attention to get $\text{A}^\prime(Q)$ at the sentence level for multiple input candidates corresponding to the same translation result. The average calculation is performed only once after all decoder layers, just before the final layer normalization, and then assign $\text{A}^\prime(Q)$ to $\text{A}_{candidate_i}(Q)$. We find that calculating only once at the final layer yields the best results. 



\vspace{-0.5cm}

\begin{align}
\text{A}^\prime(Q) = \frac{1}{n_{\text{candidate}}} \sum_{i=1}^{n_{\text{candidate}}} \text{A}_{candidate_i}(Q)
\end{align}

During the inference stage, the attention mechanism is utilized to compute the average attention score for beams that share the same candidate sequence number. This aggregation of attention scores forms a pool of candidate beams. Consider there are $m$ beams in total. Each time predicting the next token, the $n$ candidates generate $mn$ results in total. For each candidate, there are $m$ beams, and the $\text{A}_{beam_k}(Q)$ is calculated as follows. The calculation of attention occurs at the same position and training stage, after all decoder layers, just before the final layer normalization. Attention is calculated once for each token generated. Subsequently, the other calculations are the same as regular beam searches.

\vspace{-0.2cm}
\begin{align}
\text{A}_{beam_k}(Q) = \frac{1}{n_{\text{candidate}}} \sum_{i=1}^{n_{\text{candidate}}} \text{A}_{candidate_i\_{beam_k}}(Q)
\end{align}





It's worth noting that this does not add any parameters to the MT model and can be used without further training, compared with ~\cite{matusov2005integration,quan2005integrated,ma2020neural}. 
It is a lightweight method that can be readily adapted to various attention-based architectures.

\subsection{Acoustic and linguistic features fusion}

The process of converting audio recognition to text and then to machine translation further results in the loss of a large amount of acoustic information in speech. We propose a multi-candidate self-supervised learning speech machine translation model(shown in Figure 2) to enhance the accuracy of filtering correct results by leveraging language patterns from machine translation and fusing acoustic and linguistic features, following the approach utilized in previous works~\cite{lee2021direct} and utilizing HuBERT~\cite{hsu2021HuBERT} for generating target self-supervised discrete units. This choice was influenced by the superior performance demonstrated by HuBERT in various tasks such as ASR, spoken language modeling, and speech synthesis, as shown in by ~\cite{yang2021superb,lakhotia2021generative,polyak2021speech}. It outperforms other unsupervised representations, including VAE-based representations employed in ~\cite{tjandra2019speech,zhang2021uwspeech}.

We follow~\cite{lee2021direct} to use the $11^{th}$ layer of HuBERT as input, which contains richer linguistic information, and transform them into tokens with a word list of 1000 for the number of speech units using the K-means model trained on English speech. Due to the high length of the original unit, we reduced the consecutive repetitive units to a single unit. We create a new trainable vocabulary for word embeddings. After being encoded, it interacts with the decoder's cross-attention to obtain deeper language information. The calculation process is the same as that of how the output of the source text encoder computes cross-attention with the production of self-attention. The rest of the process is similar to that of the multi-candidate machine translation model.

\section{Experiments \& Results}

\subsection{Data}
For ST datasets, we use GigaST~\cite{ye2022gigast}, a large-scale pseudo speech translation (ST) corpus containing 7.5M en-zh pairs. It is created by translating the text in GigaSpeech~\cite{chen2021gigaspeech}, an English ASR corpus, into German and Chinese. The training set is translated by a robust machine translation system, and the test set is translated by humans. 

\subsection{Model setup}
\subsubsection{Cascaded system}

Our cascaded model first uses Wenet~\cite{yao2021wenet} to perform ASR on Gigaspeech with 7.4M valid en-zh pairs, and we store the speech recognition texts corresponding to the top 20 cumulative log probability value rankings at the end of beam search. The top $n$ ranked texts are input as encoders into the mBART-based machine translation model. 
We do not use any self-supervised speech models to initialize the ASR encoder. The encoder is trained from scratch following the WeNet paper. Taking inspiration from the approach proposed by ~\cite{lee2021direct}, We use the multilingual HuBERT (mHuBERT) model and K-means model to encode source speech into a vocabulary of 1000 units. The mHuBERT and the K-means models are learned from the combination of English, Spanish, and French unlabeled speech data from VoxPopuli~\cite{wang2021voxpopuli}, while we use them to encode English speech only.

\subsubsection{Multi-candidate mBART} 
The multi-candidate mBART is built upon the mBART model~\cite{liu2020multilingual}. We fine-tune the mBART model on 8 A100 GPUs. Our model configuration follows\footnotemark to use the same parameters for fine-tuning the mBART model. 

\footnotetext{https://github.com/facebookresearch/fairseq/blob/main/examples/\\mbart/README.md}

On the GigaST dataset, we compare our method with the original cascaded system and two end-to-end ST systems, including SSL-Transformer and Speech-Transformer~\cite{ye2022gigast}.



\vspace{-0.3cm}

\begin{table}[ht]
\caption{Giga-ST main results: “MC” denotes the method of multi-candidate in Section 3.1.2, “Alignment” denotes the method in Section 3.1.1, “sslS" denotes the method of fusing acoustic and linguistic features in Section 3.2}
    \centering
    \begin{tabular}{|c|c|c|}
\hline Settings & Models & BLEU score  \\
     
\hline
  (1)   &  Machine translation  & 40.2  \\

\hline \multicolumn{3}{|l|}{ End-to-End Model } \\
\hline(2)   & SpeechTransformer & 36.3  \\
   (3)   &    SSL-Transformer  & 38.0  \\
       
\hline \multicolumn{3}{|l|}{ Cascaded Model } \\
\hline 
     (4)   & Wenet + mBART  & 36.8  \\
     (5)   &    (4) + MC & 36.9 \\
   (6)   &    (4) + MC + Alignment & 37.8 \\
  (7)   &    (4) + MC + Alignment + sslS   & 38.1\\

\hline
\end{tabular}
    
    \label{tab:mustc-results}
\end{table}


       

    

\subsection{Results}
\subsubsection{Comparison with baselines}

Comparing settings (6) and (4), it shows that the Alignment and Multi-candidate methods enhance the performance of the conventional cascaded system in setting (4). When comparing settings (7), (6), and (3), it shows that fusing acoustic and linguistic features improves the performance of the conventional cascaded system in setting (4), making it comparable to the end-to-end model.


Moreover, while the transition from high-quality machine translation models to improved multi-candidate translation models requires no parameter changes, in the era of emerging and popular large language models, it is evident that our approach holds more excellent practical value and prospects than the End-to-End model. Furthermore, our model achieves a BLEU score of 37.3 after just one epoch of training, requiring only one hour on our settings. With a concise duration of fine-tuning, it surpasses the performance of former cascaded models, demonstrating the effectiveness of our approach.


\subsubsection{Ablation}
Comparing settings (5) and (6) shows that the Alignment method is crucial to make the multi-candidate method effective. Without lexical alignment, the sentence-level Multi-candidate average attention struggles to select the correct candidate words.




\subsubsection{Case study}
By employing the multi-candidate strategy, we can observe that among the top five candidates based on their scores, there exist samples that deviate from conventional human language expressions. Nevertheless, in the attention mechanism of the machine translation process, words that align more closely with human expression and convey correct semantic meaning receive greater attention. 
As for the example of ``Where the Golgi apparatus, sometimes called the Golgi body, receives them.'' The first candidate from ASR misrecognizes the pronunciation of the ``golgi apparatus'' as two non-existent words, ``golgy'' and ``golji''. However, the candidate with a BLEU score of 100 is ranked fifth. The machine translation model leverages rich text patterns through multi-candidates, allocates more attention to the correct candidate ``Golgi apparatus'' and effortlessly selects the proper translation, regarding the example of ``Recording the transaction in an immutable distributed ledger'' while the subsequent candidates include ``legend'', ``literature'', ``letter'', ``ledger'' and other alternatives.

\section{Conclusion}
Our analysis pinpoints factors contributing to error propagation in cascaded systems, such as pronunciation disparities and semantic differences. Our multi-candidate approach notably enhances speech-to-text (S2T) translation, bridging the S2T-T2T gap without altering model parameters. With enhanced ASR and MT resources, our multi-candidate method narrows the S2T-T2T divide, providing increased accuracy and efficiency, all without additional parameters or modules.

\bibliographystyle{IEEEtran}
\bibliography{IEEE-conference-template-062824}

\end{document}